\documentclass[10pt,twocolumn,letterpaper]{article}

\usepackage{iccv}              

\usepackage{times}
\usepackage{epsfig}
\usepackage{graphicx}
\usepackage{amsmath}
\usepackage{amssymb}
\usepackage{microtype}
\usepackage{graphicx}
\usepackage{subcaption}
\usepackage{booktabs} 
\usepackage{amssymb}
\usepackage{amsmath}
\usepackage{mathtools}
\usepackage{booktabs} 

\usepackage{multirow}
\usepackage{booktabs} 
\usepackage{arydshln}  

\newcommand{\fref}[1]{Fig. ~\ref{#1}}
\newcommand{\tref}[1]{Table ~\ref{#1}}

\definecolor{iccvblue}{rgb}{0.21,0.49,0.74}
\usepackage[pagebackref,breaklinks,colorlinks,allcolors=iccvblue]{hyperref}

\def\paperID{18} 
\def\confName{ICCV}
\def\confYear{2025}

\title{Artist-Created Mesh Generation from Raw Observation}

\author{Yao He, Youngjoong Kwon, Wenxiao Cai, Ehsan Adeli\\
Stanford University\\
}

\begin{document}
\maketitle
\begin{abstract}
We present an end-to-end framework for generating artist-style meshes from noisy or incomplete point clouds, such as those captured by real-world sensors like LiDAR or mobile RGB-D cameras. Artist-created meshes are crucial for commercial graphics pipelines due to their compatibility with animation and texturing tools and their efficiency in rendering. However, existing approaches often assume clean, complete inputs or rely on complex multi-stage pipelines, limiting their applicability in real-world scenarios. To address this, we propose an end-to-end method that refines the input point cloud and directly produces high-quality, artist-style meshes. At the core of our approach is a novel reformulation of 3D point cloud refinement as a 2D inpainting task, enabling the use of powerful generative models. Preliminary results on the ShapeNet dataset demonstrate the promise of our framework in producing clean, complete meshes. 

\end{abstract}

\vspace{-1em}
    
\section{Introduction}

Artist-created meshes (AMs) play a central role in 3D applications due to their compatibility with commercial graphics pipelines. These meshes offer high-quality topology using relatively few vertices, making them ideal for efficient rendering and editing. However, generating artist-style meshes from real-world sensor data remains a significant challenge.

Existing mesh generation pipelines~\cite{meshgpt, meshanything, chen2024meshanything, hao2024meshtron, deepmesh} often assume access to clean and complete 3D inputs, such as dense point clouds or high-fidelity scans. In practice, however, point clouds captured by real-world sensors—like LiDAR or mobile RGB-D cameras—are frequently sparse, noisy, or incomplete. Moreover, conventional approaches rely on complex, multi-stage processing pipelines, which are difficult to scale and prone to cascading errors.

To address these limitations, we propose an end-to-end framework that generates high-quality artist-style meshes directly from raw point cloud inputs, even when the input data is noisy or partial. At the core of our method is a novel formulation of 3D point cloud refinement as a 2D inpainting problem, enabling the use of powerful 2D generative models (\textit{i.e.}, Stable Diffusion~\cite{stablediffusion}).

Specifically, we first project the input 3D point cloud onto a spherical atlas using sphere offsetting followed by equirectangular projection. This produces a 2D point cloud atlas that retains geometric and structural information in image space. We then apply a denoising U-Net to inpaint the atlas, filling in missing regions and reducing noise. The refined atlas is subsequently mapped back to 3D and converted into a clean, complete point cloud. Finally, we pass this point cloud through a frozen feed-forward transformer~\cite{meshanything} to generate the final artist-style mesh.

Our contribution can be summarized as follows:
\begin{itemize}
    \item An end-to-end framework for generating artist-style meshes directly from raw point cloud inputs.
    \item A novel formulation of 3D point cloud refinement as a 2D inpainting problem.
    \item Promising results on the ShapeNet dataset~\cite{shapenet}, showing the potential of our approach for real-world applications.
\end{itemize}

\section{Related Work}
\begin{figure*}[t]
	\begin{center}
  \includegraphics[width=1\linewidth]{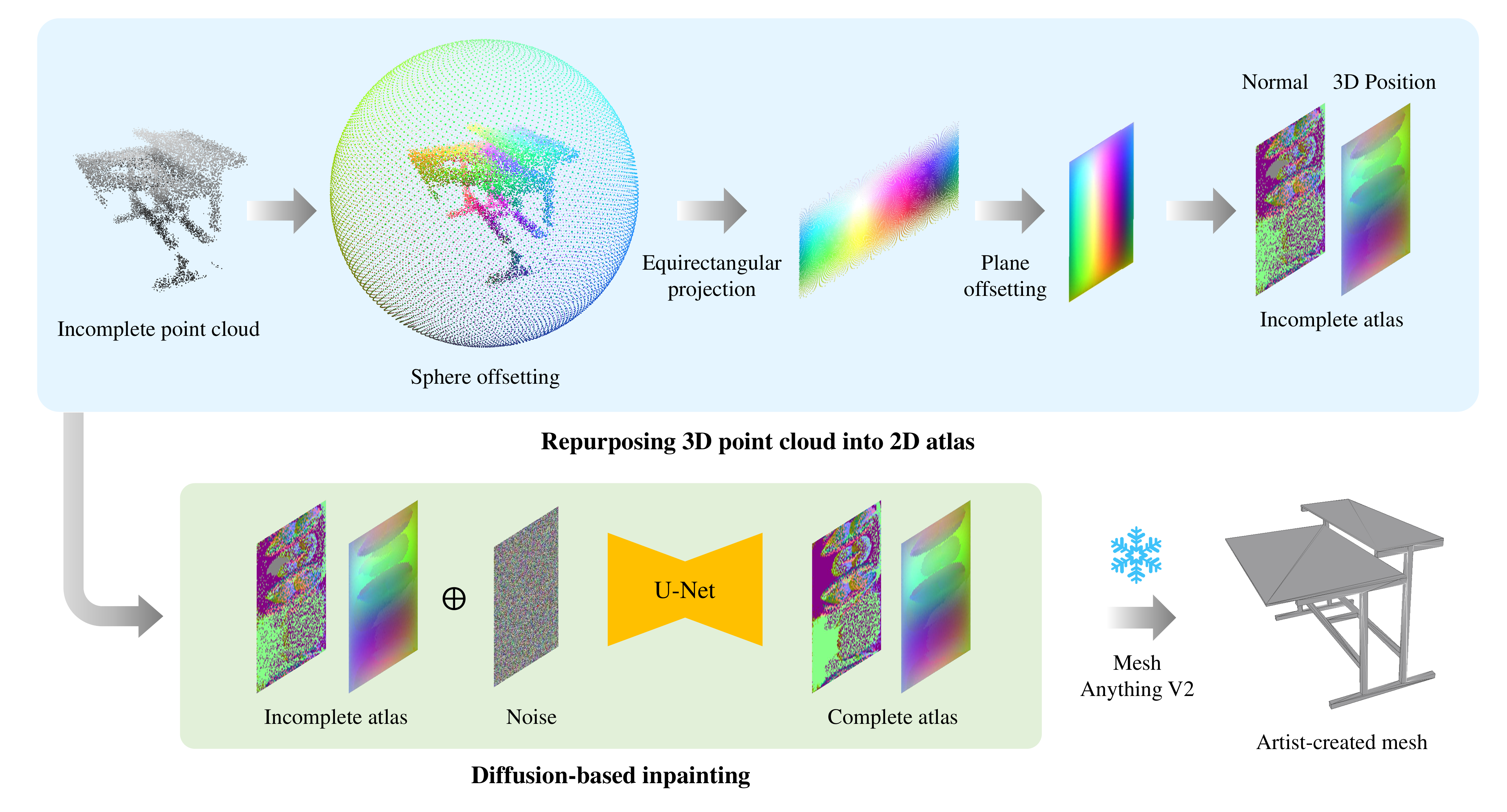}
  \end{center}
  \vspace{-1em}
  \caption{\textbf{Overview of Our Pipeline.} \textit{Top}: Representing 3D Point Cloud as 2D Atlas. For each fitting, the point cloud is first translated to the surface of a standard sphere $S$ via optimal transport, where points with the same color indicate a 1-to-1 mapping. Then, the surface points are flattened onto the 2D plane via equirectangular projection with reusable indices. We obtain Atlas by reorganizing the flattened coordinates to pixels of a dense 2D square of size $\sqrt{N}\times\sqrt{N}$. \textit{Bottom}: Training And Inference Pipeline. We fine-tune a latent diffusion model to inpaint incomplete atlases. The completed atlas is then used to reconstruct the full point cloud through inverse mapping of optimal transport. The point cloud is subsequently passed to an off-the-shelf mesh generation model to produce the final artist-created mesh.}
  \label{fig:overview}
  \vspace{-1.0em}
\end{figure*}

\subsection{Point Cloud Refinement}
The point cloud refinement task aims to recover missing geometry from partial or noisy point clouds. Early deep learning approaches~\cite{dai2017shape,han2017high,stutz2018learning,varley2017shape} leveraged 3D convolutional networks on voxelized inputs, but these were constrained by high computational and memory costs. PointNet~\cite{qi2017pointnet,qi2017pointnet++} introduced a paradigm shift by directly operating on unordered point sets, enabling more scalable processing. Building on this, PCN~\cite{yuan2018pcn} and subsequent works~\cite{liu2020morphing,wang2020cascaded,wen2022pmp,xiang2021snowflakenet} adopted a coarse-to-fine folding-based architecture to generate dense point clouds in an end-to-end fashion, conditioned on partial input.
Transformer-based methods have since improved completion quality. However, they often suffer from limited contextual understanding, especially when relying solely on 3D input. To address this, recent methods incorporate image guidance. For instance, MBVN~\cite{hu2019render4completion} employs a conditional GAN to complete missing regions in 2D space and reconstructs 3D point clouds from the completed images. ViPC~\cite{zhang2021view} and Cross-PCC~\cite{wu2023leveraging} use both the partial point cloud and a single-view image for completion. SVDFormer~\cite{zhu2023svdformer} and PCDreamer~\cite{wei2024pcdreamer} take a further step by synthesizing novel views and using them to guide point cloud generation. In particular, PCDreamer utilizes a diffusion-based prior to generate realistic multi-view images. However, due to the lack of strong multi-view consistency in diffusion models, the resulting 3D reconstructions often remain noisy.
%
%
Importantly, most prior works focus solely on completing the missing points without regard for mesh topology. As a result, meshing these outputs often leads to overly complex and unstructured surfaces. 
Our approach not only refubes the point cloud but also explicitly aims to produce artist-created meshes with clean topology, suitable for downstream use in commercial graphics pipelines.

\subsection{Artist-created Mesh Generation}
Recent works on artist-created mesh generation focus on producing meshes that efficiently and faithfully capture the underlying topology. A prominent line of research~\cite{alliegro2023polydiff,chen2024meshxl,chen2024meshanything,nash2020polygen,siddiqui2024meshgpt,weng2024pivotmesh} formulates mesh generation as a sequence modeling task, where the mesh is represented as an ordered sequence of faces or vertices. PolyGen~\cite{nash2020polygen} employs two autoregressive transformer models to generate vertex coordinates and face indices separately. MeshXL~\cite{chen2024meshxl} directly autoregresses over sequences of explicit vertex coordinates to generate raw meshes. MeshGPT~\cite{meshgpt} tokenizes geometric structures into a learned vocabulary and decodes these tokens into faces using an autoregressive model. Polydiff~\cite{alliegro2023polydiff} represents the mesh as a quantized triangle soup and applies a diffusion model to denoise it into a clean, coherent mesh.
To improve structure and quality, PivotMesh~\cite{weng2024pivotmesh} introduces pivot vertices as coarse anchors to guide mesh token generation. MeshAnything~\cite{meshanything} leverages a VQ-VAE to learn a compact mesh vocabulary and generates meshes conditioned on input shapes using an autoregressive transformer. Its successor, MeshAnythingV2~\cite{meshanythingv2}, improves tokenization efficiency by encoding each face with a single vertex. Meshtron~\cite{meshtron} proposes a scalable solution using an hourglass-style transformer architecture for mesh sequence generation.
These approaches aim to bridge the gap between raw geometric reconstructions and structured, artist-quality outputs suitable for commercial use. In our work, we align with this goal by generating artist-created meshes from incomplete or noisy point clouds, while also ensuring topological consistency and mesh simplicity through a 2D inpainting-driven design.

\section{Method}

\label{sec:proposed_method}

We formalize the task of generating a high‐fidelity artist‐created mesh from an incomplete and noisy point cloud. Let $\mathcal{P}_{\text{gt}} \subset \mathbb{R}^3$ denote the ground truth point cloud and $\mathcal{P}_i \subset \mathcal{P}_{\text{gt}}$ denote a subset of the ground truth. Our input is a noisy corrupted version of $\hat{\mathcal{P}}$, defined as:
\begin{equation}
    \hat{\mathcal{P}} = \left\{ p + n_p \mid p \in \mathcal{C}_i(\mathcal{P}_{\text{gt}}),\ n_p \sim \mathcal{N}(0, \sigma^2 I) \right\}.
\end{equation}
where $\mathcal{C}_i: \mathbb{R}^3 \to \mathbb{R}^3$ is a corruption operator on a point cloud, selecting a subset of $\mathcal{P}_{\text{gt}}$ and $\mathcal{N}_i$ represents the noise applied to the points. The resulting point cloud captures real-world scanning artifacts such as sparsity, occlusions, and sensor noise. Our desired output is a mesh $\boldsymbol{M}$ adopting the form of ordered sequence of faces, vertices and coordinates as described in \cite{hao2024meshtron}.
We adopt this representation to align with the standard sequence generation formulation for mesh generation. However, due to the noisy and corrupted nature of the input point cloud, additional steps are necessary to address these challenges.

We aim to learning the distribution $P(M\mid\hat{S})$ where $M$ represents the AM and $\hat{S}$ represents the 3D shape condition from the incomplete point cloud $\hat{P}$. 
In contrast, prior work~\cite{meshanything, chen2024meshanything, hao2024meshtron, deepmesh} focuses on learning the distribution $P(M\mid S)$ where $S$ is the 3D shape condition from the complete and noise-free point clouds. A straightforward approach is to fine-tune these models so that they adapt from $P(M \mid S)$ to $P(M\mid \hat{S})$
. However, as pointed out in ~\cite{hao2024meshtron}, these methods often fail to produce meaningful results when conditioned on incomplete or noisy 3D data. To address this challenge, we instead factorize the distribution as:
\begin{equation}
  P(M \mid \hat{S}) = P(M \mid S)\,P(S \mid \hat{S})
  \label{main_eq}
\end{equation}

This factorization suggests a two-stage approach: first, recover the complete shape condition from its corrupted counterpart by modeling $P(S \mid \hat{S})$, then, generate the AM using the standard pipeline based on $P(M \mid S)$. An overview of our method is demonstrated in \fref{fig:overview}.

\subsection{Recovering Complete Shapes}
Our key intuition is to leverage the powerful distribution modeling capabilities of diffusion models to recover the complete shape condition $S$ from the corrupted version $\hat{S}$, i.e., learning $P(S \mid \hat{S})$. While diffusion models have demonstrated strong generative performance in the 2D domain, 3D diffusion model has been limited by the scarcity of high-quality 3D data, resulting in performance gaps relative to their 2D counterparts. To overcome this limitation, we adopt recent approaches that re-purpose pre-trained 2D diffusion models for 3D object generation through atlasing \cite{xiang2025repurposing, yang2024atlas, zhang2024gaussiancube}. Our pipeline for representing 3D point clouds as 2D atlas is shown on top of \fref{fig:overview}.

Specifically, this re-purposing process begins by hypothesizing a unit sphere centered around the object, populated with $N$ 3D points $\boldsymbol{s} = \{s_i \vert s_i \in \mathbb{R}^3\}$ uniformly distributed on its surface. Each point in the input point cloud is then mapped onto the sphere surface via Optimal Transport (O. T.) \cite{burkard1999linear}. Once positioned, an equirectangular projection is applied to convert the spherical coordinates to flattened 2D coordinates ${m_i \in \mathbb{R}^2}$. A second O. T. step maps these projected coordinates $m_i$ onto the vertices $n_i \in \mathbb{R}^2$  of a regular $\sqrt{N} \times\sqrt{N}$ 2D grid. While prior methods apply Gaussian atlases to encode features such as 3D location, color, opacity, scale, and rotation, our point cloud setting includes only 3D locations and normals. Therefore, the final output is an atlas $\mathbf{X} \in \mathbb{R}^{\sqrt{N} \times\sqrt{N} \times (\mid \mid \boldsymbol{x} - \boldsymbol{s} \mid \mid + \mid \mid \boldsymbol{n} \mid \mid )}$ where $\boldsymbol{x} - \boldsymbol{s}$ and $\boldsymbol{n}$ denote the offset 3D coordinates and normals, respectively. 

Upon obtaining the atlas map, we fine-tune the pre-trained Latent Diffusion (\textit{i.e.}, Stable Diffusion~\cite{stablediffusion}) to inpaint missing areas of the given incomplete atlas. Following \cite{xiang2025repurposing}, our inpaint U-Net $\mathcal{U}$ directly operates on the atlas without involving VAE. At each diffusion step $t$, the inpaint network $\mathcal{U}$ predicts the noise $\epsilon$ added to the ground-truth atlas $\mathbf{X}$, conditioned on the input incomplete atlas $\mathbf{\hat{X}}$. Specifically, we condition the generation on the incomplete atlas by adding it to the noisy image (\textit{i.e.}, $\mathbf{X}_{t} + \mathbf{\hat{X}}$). Although we used the simple conditioning for the initial exploration, in the future work, more sophisticated conditioning (\textit{e.g.}, self-attention) can be employed. The training objective minimizes the difference between the predicted and the ground-truth noise: $\mathcal{L} = \mathbb{E} \left[ \| \epsilon - \mathcal{U}(\mathbf{X}_{t} + \mathbf{\hat{X}}, i) \|^2 \right]$, where $\mathbf{X}_{t} = \alpha_t \mathbf{X} + \sigma_t \epsilon$.
Here, $\mathbf{X}$ is the ground-truth latent, $\epsilon \sim \mathcal{N}(0, I)$ represents Gaussian noise, and $\alpha_{i}$ and $\sigma_{i}$ are diffusion parameters that define the noise level at diffusion timestep $i$.


During inference time, we use $T=20$ denoising steps to inpaint the corrupted atlas. The completed atlas is then transformed back into a 3D point cloud representation from the bijective O. T. mapping, thereby closing the loop for learning the distribution $P(S\mid\hat{S})$.

\subsection{Recovering Meshes}
Given an estimate of $P(S\mid\hat{S})$, we can approximate $P(M\mid\hat{S})$ by first computing $P(M \mid S)$ as described in Eq. \ref{main_eq}. At this stage, the availability of a complete point cloud allows us to employ an off-the-shelf mesh generation model. Specifically, we use MeshAnything V2~\cite{chen2024meshanything} to estimate $P(M \mid S)$. It formulates mesh generation as an auto-regressive token prediction problem on a learned mesh vocabulary. Given a fully completed shape $S$ condition, i.e., the encoded point cloud, it infers the conditional distribution
\begin{equation}
    P(M \mid S) = \prod_{t=1}^{T} P\bigl(m_t \mid m_{1:t-1},\,S\bigr)
\end{equation}
where each token $m_t$ represents either a vertex coordinate quantized to a fixed grid or a face index triplet. 

\begin{figure*}[t]
	\begin{center}
  \includegraphics[width=1\linewidth]{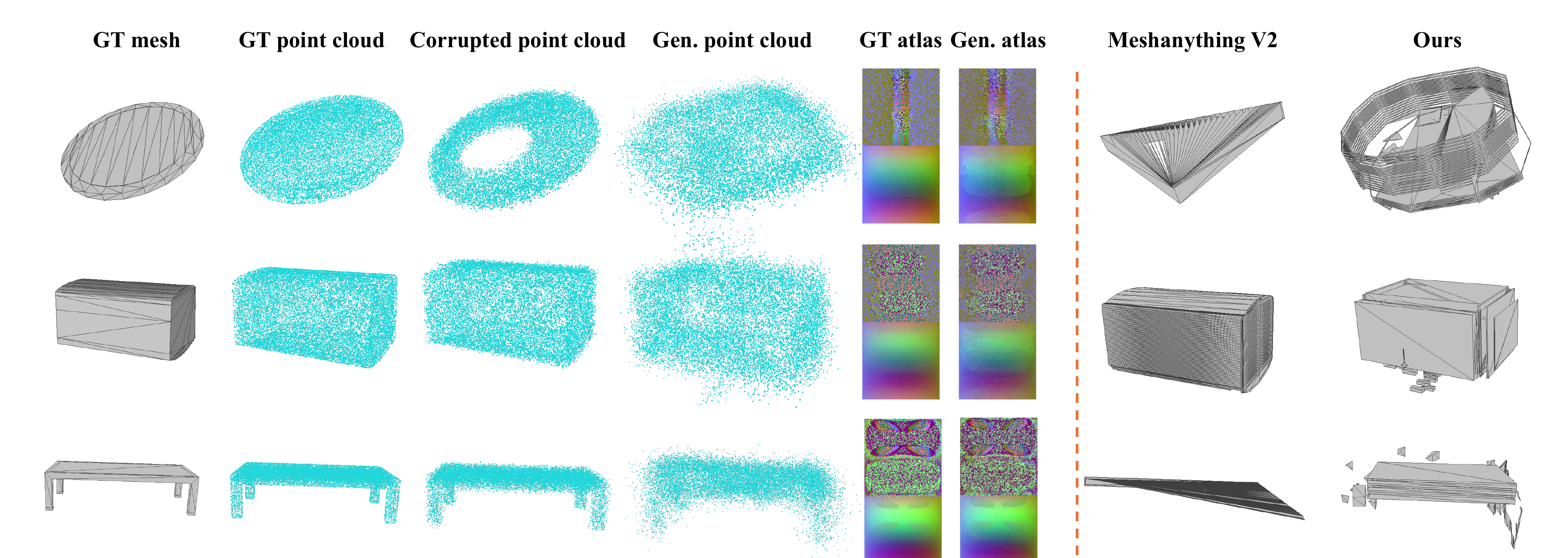}
  	\end{center}
  \caption{\textbf{Qualitative Results.} We compare the meshes generated by our method and MeshAnything V2. The figure also includes the ground-truth mesh, point cloud sampled from the ground-truth meshes, corrupted input point cloud, ground-truth Atlas maps, and their generated counterparts. The results demonstrate that our approach can reconstruct key geometric features where MeshAnything V2 fails, while also highlighting several limitations of our model.}
  \label{fig:qualitative_results}
\end{figure*}
\begin{table*}[t]
  \centering
  \caption{Comparison of our method and MeshAnything V2 on the noisy and cropped ShapeNet dataset.}
  \label{tab:main}
  \begin{tabular}{lccccccc}
    \toprule
    Method & CD$\downarrow$ (×10$^{-2}$) & ECD$\downarrow$ (×10$^{-2}$) & NC$\uparrow$ & \#V$\downarrow$ & \#F$\downarrow$ & V\_Ratio$\downarrow$ & F\_Ratio$\downarrow$  \\
    \midrule
    Ours            & 0.913          & \textbf{1.283} & 0.344          & \textbf{381.2} & \textbf{635.0} & \textbf{3.505} & \textbf{1.599}  \\
    MeshAnything V2 & \textbf{0.909} & 1.422          & \textbf{0.393} & 527.4          & 998.1 & 4.705 & 2.398  \\
    \bottomrule
  \end{tabular}
\end{table*}
\section{Experiments}
\label{sec:exp}

In this section, we benchmark and evaluate the effectiveness of our proposed method on the test set from our dataset described above. MeshAnything~V2~\cite{chen2024meshanything} serves as our primary baseline. 
To quantitatively assess mesh quality, we adopt the following standard metrics setting as in \cite{chen2022ndc}:
Chamfer Distance ($CD$),
Edge Chamfer Distance ($ECD$),
Normal Consistency ($NC$),
Vertex and Face Counts ($\#V$, $\#F$),
Vertex/Face Ratios ($V\_Ratio$, $F\_Ratio$).



\subsection{Results \& Discussion}
We present the evaluation results of our method and MeshAnything V2 in Table~\ref{tab:main}. ~\fref{fig:qualitative_results} shows qualitative examples of the generated meshes.

Overall, the evaluation results in ~\tref{tab:main} show that our approach achieves performance comparable to, and in some cases surpassing, baseline, demonstrating its effectiveness under noisy point cloud input conditions. In particular, our method achieves a similar $CD$, and a better $ECD$, suggesting comparable mesh quality. However, it performs worse in NC, indicating challenges in accurately reconstructing surface normals. This may be attributed to the standard diffusion process not enforcing unit-length constraints, leading to noisier outputs. This insight suggests a future improvement in the training pipeline.

We observe that our method achieves lower values in the metrics $\#V$, $\#F$, $V\_Ratio$, $F\_Ratio$ compared to MeshAnything V2. While lower values in these metrics typically indicate higher ability to represent complex topology using fewer vertices and faces, this does not apply in our case. A sanity check of the generated results reveals that many generated meshes exhibit missing parts. Thus, the reduced values more likely reflect a deficiency in preserving geometric completeness. We provide further hypotheses for this behavior in the supplementary materials.

\section{Conclusions}

We presented an end-to-end framework for generating artist-style meshes from noisy or incomplete point clouds, addressing key limitations of existing methods that assume clean inputs or rely on multi-stage pipelines. By reformulating the 3D point cloud refinement problem as 2D inpainting, our approach leverages powerful 2D generative models to recover clean and complete geometry. Preliminary results demonstrate the potential of our method to produce high-quality meshes from real-world-like inputs.

\section*{Acknowledgment}
This work was partially supported by Panasonic Holdings Corporation, UST, and the Stanford Human-Centered AI (HAI) Google Cloud Credits. 

{
    \small
    \bibliographystyle{ieeenat_fullname}
    \bibliography{main}
}
\clearpage
\clearpage
\setcounter{page}{1}
\maketitlesupplementary
\appendix
\section{Dataset}
\label{sec:dataset}

Our initial dataset for training and evaluation is constructed from ShapeNet for its high data quality within limited categories. We intend to use it as a clean starting point, then progressively fuse meshes from the other repositories. We follow \cite{chen2024meshanything} to filter out meshes with more than 800 faces and manually discarded low quality meshes by checking geometric integrity, annotation completeness, and category balance. This produces roughly 5k meshes. 

The point clouds generated thus far are high-quality, free from corruption and noise. However, to simulate the challenges posed by real-world lidar data—namely, occlusion and measurement noise—we introduce a synthetic corruption procedure. Specifically, we randomly crop a portion of the point cloud to mimic occlusion and add Gaussian noise to the remaining points to simulate sensor inaccuracies. Our random cropping consists of two strategies: 1) randomly select $20\%$ to crop, and 2) crop $20\%$ of the points around a center region. While this procedure approximates real-world conditions, it may not fully capture the true distribution of lidar artifacts, suggesting a direction for future improvement, such as using a visibility mask to crop out points. After obtaining the point cloud, we extract its corresponding atlas, i.e., position map and normal map, for fine-tuning the Latent Diffusion.

\begin{figure*}[t]
	\begin{center}
  \includegraphics[width=1\linewidth]{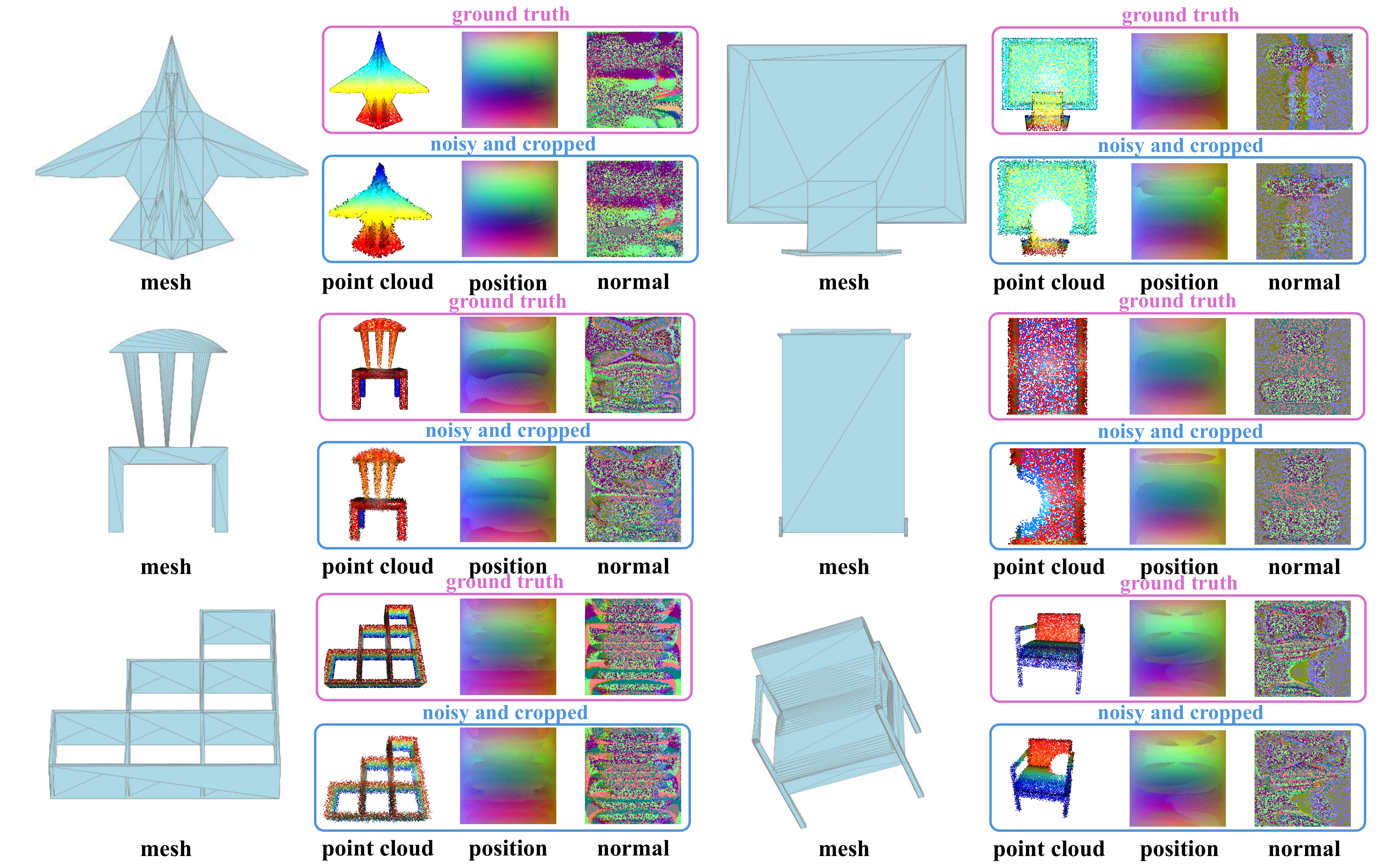}
  	\end{center}
  \caption{\textbf{Overview of Dataset}. Each data point from our processed ShapeNet includes a ground-truth mesh, ground-truth point cloud, ground-truth atlas, and their noisy and cropped counterparts. The dataset comprises approximately 5.5k samples.}
  \label{fig:dataset}
\end{figure*}

Our dataset contains approximately 5.5k samples, split into $90\%$ for training and $10\%$ for testing, following the convention in \cite{meshanything}. Each sample consists of a clean point cloud, a synthetically corrupted counterpart, and their corresponding atlases. 
Representative examples are shown in Fig. \ref{fig:dataset}.

\section{Implementation Details}
we set the total number of 3D points per atlas to $N = 16384 = 128 \times 128$. This choice of parameters balances computational efficiency for O. T. with the input resolution required for high-quality mesh generation. For point clouds with fewer than $N$ points, we pad with holes (zero positions and normals). For point cloud with more than $N$ points, we apply Farthest Point Sampling (FPS) \cite{moenning2003fast} to downsample.

The inpaint U-Net $\mathcal{U}$ is trained for 100,000 iterations with a learning rate of $1\times 10^{-5}$, using a batch size of 4 on a single NVIDIA L40S GPU. We use Adam optimizer to train the diffusion model.
The total training time is approximately 24 hours. We choose 20 denoising steps to balance the tradeoff between efficiency and generation quality during inference.
\section{Limitations And Future Works}
\subsection{Additional Discussions}
\noindent \textbf{Discussion on generated meshes}: The qualitative results in \fref{fig:qualitative_results} show that, given noisy point cloud input, our method is able to capture certain topological and geometric structures that MeshAnything V2 fails to recover, demonstrating the effectiveness of our approach. However, our model also exhibits limitations in preserving fine-grained geometry, such as the legs of a table. This is primarily due to noise in the reconstructed point clouds, which adversely affects the performance of the downstream mesh generation model.

\noindent \textbf{Discussion on generated atlas maps and point clouds}: The comparison between the generated and ground-truth atlas maps shows that our fine-tuned diffusion model can produce visually similar results when conditioned on corrupted inputs, suggesting that it captures key features of the underlying atlas map distribution. Furthermore, the reconstructed point cloud demonstrates the model’s ability to recover geometric structure through inpainting, such as filling in missing regions like the hole in a plate.

Despite promising results, our model exhibits notable limitations. A key issue is that the reconstructed point clouds reveal significant noise in the generated atlases compared to the ground truth.  This is likely due to our conditioning strategy, where we simply concatenate the input with noise; the denoising process may be insufficient to fully recover clean data, leading to noisy point cloud reconstructions. Such noisy point cloud reconstructions pose challenges for MeshAnything V2~\cite{meshanythingv2} in capturing fine-grained geometric details and introduce ambiguity in surface prediction. As a result, the MeshAnything V2 pipeline tends to prune these uncertain vertices, leading to lower vertex and face counts in the final mesh.

The findings above suggest several directions for improving our methodology. One is to explore more effective conditioning mechanisms that enable the model to better preserve fine-grained structures. Additionally, incorporating a reconstruction loss—alongside the standard diffusion loss, as proposed in \cite{xiang2025repurposing}—could further encourage the generation of more accurate atlas maps.
\subsection{Future Works}
Our initial results validate the soundness of the proposed approach and point to several directions for future work. First, we plan to expand the dataset to better reflect diverse real-world distributions. We also aim to explore alternative corruption methods, such as estimating point cloud visibility maps to more accurately simulate lidar scan artifacts. To improve atlas quality, we will investigate more effective conditioning strategies and incorporate additional loss terms. Finally, we propose bypassing the point cloud reconstruction step by directly encoding the atlas and fine-tuning the mesh generation pipeline. This would mitigate error propagation introduced by the optimal transport step during point cloud reconstruction.
\end{document}